\documentclass[sigconf]{acmart}

\usepackage{booktabs} % For formal tables

\setcopyright{none}
\begin{document}
\title{Emotion Recognition in the Wild using \\ Deep Neural Networks and Bayesian Classifiers}

\author{Luca Surace}
\affiliation{%
  \institution{University of Calabria - DeMACS}
  \streetaddress{Via Pietro Bucci}
  \city{Rende (CS)} 
  \state{Italy} 
}
\email{lucasurace11@gmail.com}

\author{Massimiliano Patacchiola}
\affiliation{%
  \institution{Plymouth University - CRNS}
  \streetaddress{Portland Square PL4 8AA}
  \city{Plymouth} 
  \state{United Kingdom} 
}
\email{massimiliano.patacchiola@plymouth.ac.uk}

\author{Elena Battini S\"onmez}
\affiliation{%
  \institution{Istanbul Bilgi University - DCE}
  \streetaddress{Eski Silahtarağa Elektrik Santralı 
Kazım Karabekir Cad. No: 2/13 
34060 Eyüp }
  \city{Istanbul} 
  \state{Turkey} 
}
\email{ebsonmez@bilgi.edu.tr}

\author{William Spataro}
\affiliation{%
  \institution{University of Calabria - DeMACS}
  \streetaddress{Via Pietro Bucci}
  \city{Rende (CS)} 
  \state{Italy} 
}
\email{william.spataro@unical.it}

\author{Angelo Cangelosi}
\affiliation{%
  \institution{Plymouth University - CRNS}
  \streetaddress{Portland Square PL4 8AA}
  \city{Plymouth} 
  \state{United Kingdom} 
}
\email{angelo.cangelosi@plymouth.ac.uk}

\renewcommand{\shortauthors}{L. Surace et al.}

\begin{abstract}
Group emotion recognition in the wild is a challenging problem, due to the unstructured environments in which everyday life pictures are taken. Some of the obstacles for an effective classification are occlusions, variable lighting conditions, and image quality. In this work we present a solution based on a novel combination of deep neural networks and Bayesian classifiers. The neural network works on a bottom-up approach, analyzing emotions expressed by isolated faces. The Bayesian classifier estimates a global emotion integrating top-down features obtained through a scene descriptor. In order to validate the system we tested the framework on the dataset released for the Emotion Recognition in the Wild Challenge 2017. Our method achieved an accuracy of $64.68\%$ on the test set, significantly outperforming the $53.62\%$ competition baseline.
\end{abstract}


%\ccsdesc[500]{Computer methodologies~Activity recognition and understanding}


\settopmatter{printacmref=false} % Removes citation information below abstract
\renewcommand\footnotetextcopyrightpermission[1]{} % removes footnote with conference information in first column
\pagestyle{plain} % removes running headers

\maketitle

\section{Introduction}

Automatic emotion recognition has recently become an important research field, due to new possible applications in social media, marketing \cite{mcduff2014automatic}, public safety \cite{clavel2008fear}, and human-computer interaction \cite{cowie2001emotion}. Emotion recognition is generally achieved through the analysis of facial muscles movements, often called action units. After isolating the face of the subject, it is possible to assign it an emotion using action units analysis. Despite noteworthy results in structured condition this approach becomes unfeasible in unstructured environments where multiple factors (e.g. occlusions, variable lighting conditions, image quality, etc.) may affect recognition.

During the years there have been many attempts to build robust methods.
For example, in \cite{orrite2009hog} the authors used histogram of gradients in order to address the problem of human emotion identification from still pictures taken in 
semi-controlled environments. In \cite{moore2011local} the influence of multiple factors (pose, resolution, global and local features) on different facial expressions was investigated. The authors used an appearance based approach dividing the images into sub-blocks and then used support vector machines to learn pose dependent facial expressions. Deep neural networks have been used in \cite{ng2015deep}. The authors started from a network pre-trained on the generic ImageNet dataset, and performing supervised fine-tuning in a two-stage process. This cascading fine-tuning achieved better results compared to a single stage fine-tuning. A significant contribution to the use of deep neural networks for emotion recognition was given in \cite{lopes2017facial} and \cite{sonmez2017convolutional}. The authors demonstrated the strength of this approach achieving state-of-art performances on the CK+ dataset \cite{lucey2010extended}. The classification of the emotion of a group of people is a different task. Previous research \cite{dhall2015automatic} focused on the development of two parallel approaches for measuring happiness level: top-down and bottom-up. One of the first articles which considered the structure of the scene as a whole is presented in \cite{gallagher2009understanding}. The authors showed that the structure of the group provides meaningful context for reasoning about the individuals and they considered the group structure from both the local and the global point of view. A complete review of all the methods is out of the scope and we refer the reader to a recent survey \cite{sariyanidi2015automatic}.

Taking into account past literature we propose a method which integrates both global and local information. A bottom-up module isolates the human faces which are present in the picture and  
gives them as input to a pre-trained Convolutional Neural Network (CNN). At the same time a top-down module finds the label associated with the scene and passes them to a Bayesian Network (BN) which estimates the posterior probabilities of each class. The output of the system is the probability of the image to belongs to three different classes: positive, neutral, and negative. Experiments were conducted on different architectures, achieving the best results with a pipeline that redirects the output of the CNN to the Bayesian classifier.
We tested the system on the dataset released for the Emotion Recognition in the Wild Challenge 2017 (EmotiW) \cite{dhall2012collecting, dhall2015more, dhall2017individual}, obtaining an accuracy of $67.75\%$ on the validation set, and $64.68\%$ on the test, outperforming the baseline of the competition \cite{dhall2017individual}.

\section{Proposed method}
Our method is based on the idea that the group emotion can be inferred using both top-down \cite{cerekovic2016deep} and bottom-up \cite{dhall2015more} approaches. The former considers the scene context, such as background, clothes, location, etc. The latter estimates the face expressions of each person in the group. We can summarize the pipeline for the bottom-up module in three steps:

\begin{enumerate}
    \item Face detection
    \item Face pre-processing
    \item CNN forward pass
\end{enumerate}

The first step is the face detection, which has been obtained through a commercial library \footnote{Google Vision API}. This step returns a list of frames containing isolated human faces. In the second step the faces are cropped, scaled and normalized. Finally in the third step the cropped faces are given as input to a pre-trained CNN and the output for each class are estimated through a forward pass. In parallel it is possible to run the top-down module which consists of three steps:

\begin{enumerate}
    \item Acquiring the scene descriptors
    \item Set evidences in the BN
    \item Estimate the posterior distribution of the BN
\end{enumerate}

\begin{figure}[ht]
\centering
\includegraphics[width=0.47\textwidth]{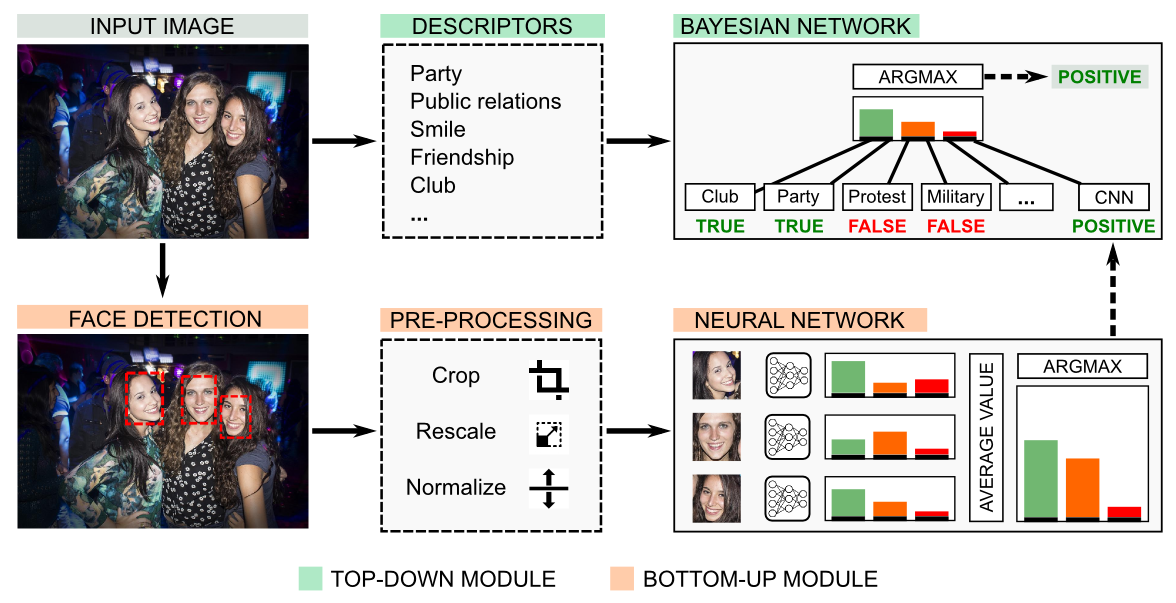}
\caption{Overview of the proposed system. In the top-down module (light-green) the scene descriptor returns a list of labels used by the BN. In the bottom-up module (light-orange) the face detector isolates the faces which are given as input to the deep neural network after a pre-processing phase. \label{fig:architecture_overview}}
\end{figure}

In the first step the top-down module estimates the content of an image returning descriptors which may be context-specific (e.g. party, protest, festival, etc.) and group-specific (e.g. team, military, institution, etc.). This step is achieved through the same previously adopted library, but it can be easily replaced with a state-of-the-art algorithm \cite{wang2015unsupervised}. In the second step, the descriptors are considered as observed variables and the corresponding nodes in the Bayesian network are set as evidence. In the third and last step the posterior distribution of the root node in the BN is estimated using the belief propagation algorithm \cite{pearl2014probabilistic}. In the next sections, details of the proposed methodology are reported, together with an overview of the entire system (Figure \ref{fig:architecture_overview}).

%SECTION - Bottom-up
\subsection{Bottom-up module}
The bottom-up module uses a CNN to estimate the emotion of isolated human faces. Deep neural networks recently achieved outstanding performances in a variety of tasks such as speech recognition \cite{hinton2012deep, graves2013speech}, and head pose estimation \cite{patacchiola2017head}.
We trained several networks having multiple architectures. The best results have been achieved with a variant of AlexNet \cite{NIPS2012_4824}. The input to the network is a color image of size 64$\times$64 pixels which is passed through 7 layers: 3 convolutional layers, 2 sub-sampling layers, 1 fully connected layer and the final output layer.
The first convolution layer produces 64 feature maps, applying a convolutions kernel of size 11 $\times$ 11. The second convolution layer generates 128 feature maps, using a convolutional kernel of size 5 $\times$ 5. The third layer produces 256 feature maps via a convolutional kernel of size 3 $\times$ 3. The sub-sampling layers use max-pooling (kernel size 3 $\times$  3) to reduce the image in half. The result of the third convolution is given as input to a fully connected layer (512 units). Finally, the network has three output units representing the three emotions: positive, neutral, negative. The first two convolutional layers are normalized with local response normalization \cite{NIPS2012_4824}. A rectified linear unit activation function is applied to each convolutional layer and to the first fully connected layer. 
For the training we used an adaptive gradient method, RMSProp \cite{Tieleman2012}, and the balanced batches technique proposed in \cite{sonmez2017convolutional}. As a loss function we used the softmax cross entropy \cite{shore1981properties} between the target $t$ and the estimated value $o$, defined as follows:

\begin{equation}
\label{eq:loss_function}
J(T, O) = - \frac{1}{N} \sum_{n=1}^{N} \Big[ t_{n} \ln{(o_{n})} + (1-t_{n}) \ln(1-o_{n}) \Big]
\end{equation}

where $N$ is the size of the batch, $T = \{t_{1}, ..., t_{N} \}$ is the set of target values, $O = \{o_{1}, ..., o_{N} \}$ is the set of output values.
Once the network has been trained it is possible to estimate the average group emotion for the faces present in the image. First of all, we averaged the predictions resulting from a forward pass on all the input faces, similarly to \cite{vonikakis2016group}. Secondly, we returned the class corresponding to the higher value. We can summarize the two steps in a single equation:

\begin{equation}
\hat{o} =  \underset{}{\operatorname{argmax}} \ \Bigg( \frac{\sum_{k=1}^{K} \sigma(\mathbf{o}_{k}) }{K} \Bigg)
\end{equation}

where $K$ is the total number of faces, $\mathbf{o}$ is a three dimensional vector representing the output of the network, and $\sigma$ is the softmax function.
The resulting scalar $\hat{o}$ represents the index of the class which better represents the scene emotion, based on all the faces that has been found in the image. This method can be extremely effective, but is has some drawbacks. First of all, it requires to identify at least one face per image. Secondly, the faces should be a good predictor of the overall emotion, which is not always the case. To compensate this source of error we used a top-down module which helps to describe the scene.

%SECTION: Top-down
\subsection{Top-down module}
\begin{figure}[ht]
\centering
\includegraphics[width=0.47\textwidth]{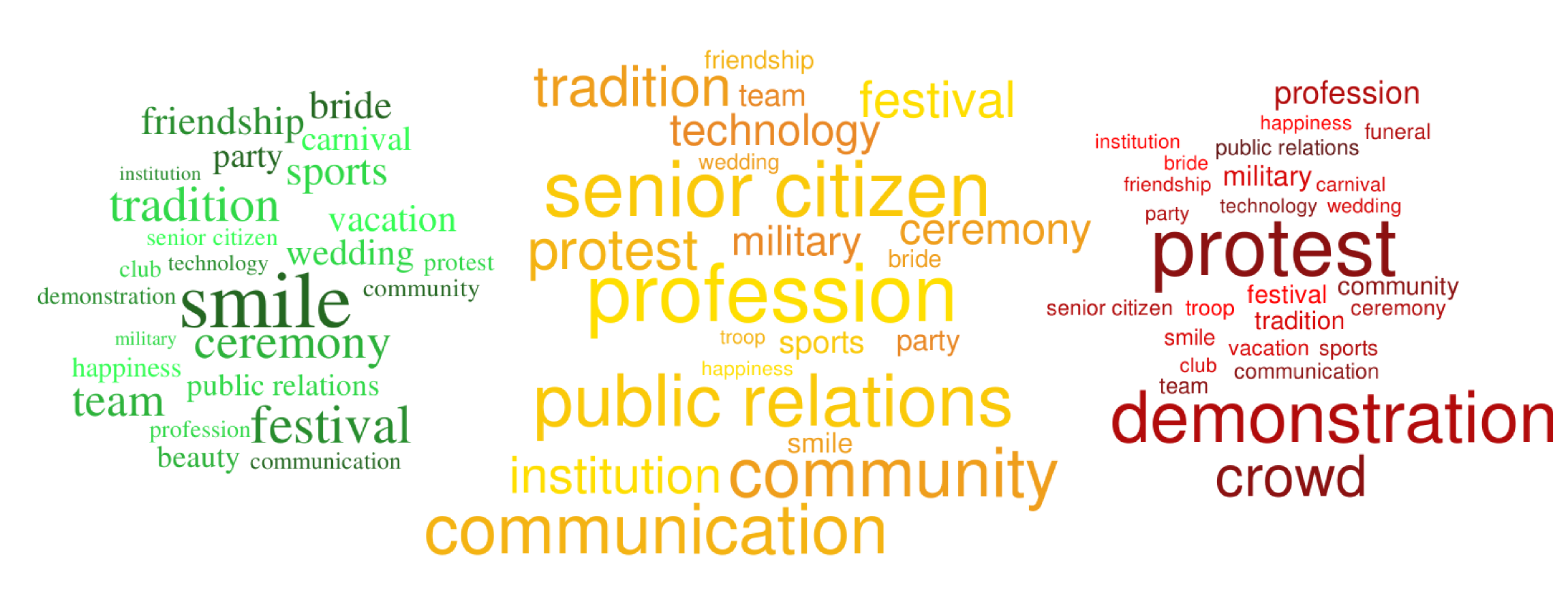}
\caption{Overview of the descriptors for each one of the three classes: positive (left-green), neutral (center-orange), negative (right-red). Words which have a larger font size, recur more frequently. \label{fig:fig_word_cloud_descriptors_pos_neu_neg}}
\end{figure}

Previous literature shows that global scene information is very useful for group emotion classification \cite{dhall2015more}. In particular, scenarios focusing only on small details can easily lead to miss-classification errors. This is why in the top-down module we used a whole-scene descriptor. In this section we describe the procedure for collecting the descriptors and how they have been integrated in the BN.
In a preliminary phase we collected meaningful descriptors for each image contained in the training set. Subsequently, we built an histogram of descriptors for the three emotions. We found a total of 812 scene descriptors, appearing with a certain frequency in the dataset. The descriptors are represented in a word cloud in Figure \ref{fig:fig_word_cloud_descriptors_pos_neu_neg}. It is possible to see how descriptors such as smile, friendship, festival, etc. recur with an high frequency in positive images and less frequently in the neutral and negative groups.

The descriptors obtained through the preliminary phase have been used as dependent nodes in a BN.
BNs represent a valid formalism to model probabilistic relationships between several random variables and their conditional dependencies via a directed acyclic graph. They have been used in different applications such as medical diagnosis \cite{onisko1999bayesian}, and cognitive modeling \cite{patacchiola2016developmental} (for a review see \cite{Friedman1997}).
In this work we started from the assumption of independence between each pair of descriptors. This assumption is an oversimplification because does not capture relationships between different features. However, from a practical point of view it works extremely well, and it is applied in many state-of-the-art classifiers. From the mathematical point of view we want to estimate $\hat{y}$, the class having the higher probability given the observed descriptors:

\begin{equation}
\label{eq:eq_argmax_posterior_distribution}
\hat{y} = \underset{y}{\operatorname{argmax}} \ P(y \vert x_{1}, x_{2}, ... , x_{N})
\end{equation}

The posterior distribution associated with the root node $y$ is proportional to the product of the prior $P(y)$ and the likelihood $P(x_{i} \vert y)$ of each one of the $N$ dependent variables $x$, as follows:

\begin{equation}
\label{eq:eq_posterior_distribution}
P(y \vert x_{1}, x_{2}, ... , x_{N}) = P(y) \prod_{i=1}^{N} P(x_{i} \vert y)
\end{equation}

In our particular case the root node is a multinomial probability distribution which models the three possible outcomes: positive, neutral and negative. For each descriptor there is a dependent variable, which is modeled with a Bernoulli distribution: true (present), false (absent). The main point for obtaining $\hat{y}$ is to find the conditional probabilities $P(x_{i} \vert y)$. Since for the training set we know the emotion associated with every image, we can use Maximum Likelihood Estimation (MLE) \cite{scholz1985maximum} to find the conditional probability distributions for each node. For instance, naming $N_{t}^{+}$ the number of time a specific descriptor $x_{i}$ has been counted in conjunction with positive images, and $N_{f}^{+}$ the number of time that descriptor has not been counted, we can estimate the conditional probability as follows:

\begin{equation}
\label{eq:eq_maximum_likelihood_estimation}
\begin{aligned}
P(x_{i} = \text{true} \vert y= \text{positive}) &= \frac{N_{t}^{+}}{N_{t}^{+}+N_{f}^{+}} \\
P(x_{i} = \text{false} \vert y= \text{positive}) &= \frac{N_{f}^{+}}{N_{t}^{+}+N_{f}^{+}}
\end{aligned}
\end{equation}

For the same descriptor we can also estimate the probability $P(x_{i} \vert y= \text{neutral})$ of being associated with a neutral emotion, and $P(x_{i} \vert y= \text{negative})$ the probability of being associated with a negative emotion. Using these probabilities we can estimate the Bernoulli distributions associated with every descriptor and build the corresponding conditional probability tables in the BN.

\subsection{Integration}
There are different ways the results of the two modules can be combined. For instance, considering the two modules as a committee of experts we can use ensemble averaging to reduce the error of the models. Another possibility is to redirect the value obtained by the bottom-up module as input to the BN in the top layer. After a preliminary research we decided to adopt the second solution.

Going back to Equation \ref{eq:eq_argmax_posterior_distribution} and \ref{eq:eq_posterior_distribution}, we can hypothesize the presence of an additional input feature $x_{N+1}$ having as prior a three-categorical multinomial distribution (positive, neutral, negative). In order to integrate the new node in the BN it is necessary to estimate the conditional probability table for $P(x_{N+1} \vert y)$.
Similarly to Equation \ref{eq:eq_maximum_likelihood_estimation} it is possible to use MLE to find the conditional distributions for the dependent node. In the particular case considered here the conditional distribution is represented by the confusion matrix obtained testing the network on the dataset.

\section{Experiments}
In this section we describe the methodology followed during the training phase and the results achieved. We evaluate the proposed method on the GAF database \cite{dhall2015more}. This database consists of 6470 total images, which have been divided in 3633 images for the training set, 2065 for the validation set, and 772 for the test set. The dataset contains images obtained from social networks captured during social events. These images can be from positive social events (marriages, parties, etc.), neutral event (meetings, convocations, etc.), or negative events (funeral, protests, etc).
The baseline score for this dataset has been obtained using the CENTRIST \cite{wu2011centrist} approach and support vector regression. CENTRIST is a scene descriptor and is computed on the whole image. It takes into consideration both the bottom-up and top-down attributes. The classification accuracy is used as the metric in the challenge. The model baseline achieved $52.97\%$ on the Validation set and $53.62\%$ on the test set.

\subsection{Methods}
The CNNs have been trained on the dataset available with the challenge. For each one of the images in the training set we isolated the faces and performed different pre-processing operations. First of all the faces have been cropped and re-scaled to 64 $\times$ 64 pixels. Then a min-max normalization has been applied. During the training we randomly selected a balanced batch of 63 images (21 for each emotion) and performed gradient descent for 1500 epochs.
As optimizer we use the RMSProp \cite{Tieleman2012}, which has been selected after a preliminary comparison between other methods such as Adagrad \cite{duchi2011adaptive} and Adam \cite{kingma2014adam}. A learning rate $\alpha= 10^{-3}$, a decay of 0.9, and $\epsilon= 10^{-10}$ were adopted. The weights of the CNNs have been initialized using the Xavier initialization method \cite{glorot2010understanding}. 
We used dropout \cite{srivastava2014dropout}  with 0.5 probability in between the internal layers in order to prevent overfitting. 

We implemented the algorithm in Python using the Tensorflow library \cite{Tensorflow2015} for the CNNs training, and OpenCV \cite{Opencv2015} for the pre-processing operations on the images.
Experiments were carried out on a workstation having 16 cores processor, 32 GB of RAM, and the NVIDIA Tesla K40 graphical processing unit. On this hardware the training of the CNN took approximately 8 minutes. The evaluation on the validation set (2065 images) using the whole pipeline took 325 minutes.

\subsection{Results}
\begin{figure}[ht]
\centering
\includegraphics[width=0.47\textwidth]{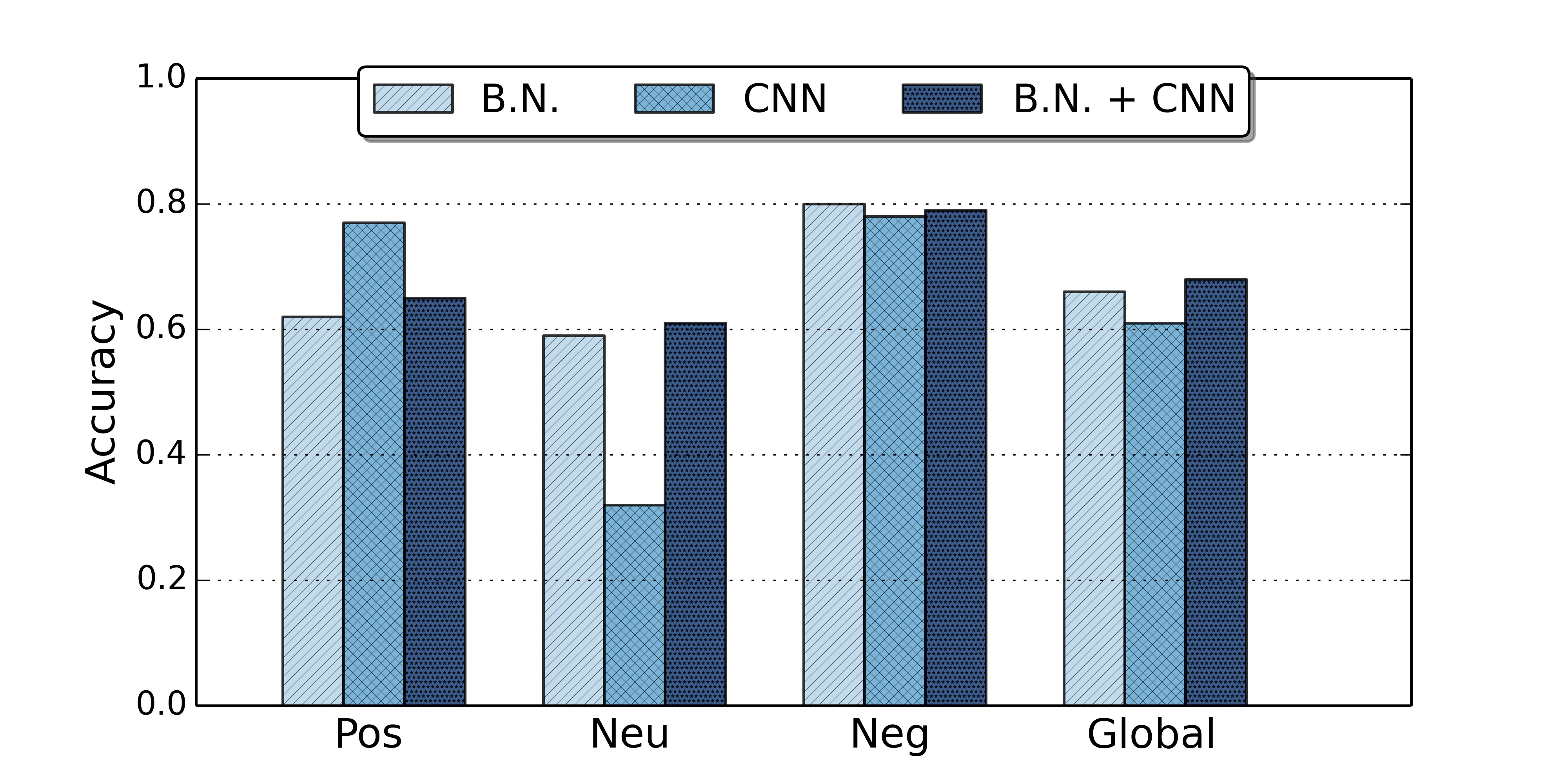}
\caption{Validation accuracy comparison for the stand-alone solutions and the complete system (BN and CNN). \label{fig:accuracy}}
\end{figure}

We obtained the best performance with the ensemble method, which lead to an accuracy of $67.75\%$ on the validation set and $64.68\%$ on the test set. Those results are significantly higher than the challenge baseline accuracy of $52.97\%$ (validation) and $53.62\%$ (test). 
Comparative results between the BN-only, CNN-only and ensemble approaches are showed in Figures \ref{fig:accuracy}. The ensemble method outperformed the results obtained using the isolated modules, supporting previous work that demonstrated how an ensemble can improve the performance of emotion recognition systems \cite{glodek2011multiple}.

\begin{figure}[ht]
\centering
\includegraphics[width=0.47\textwidth]{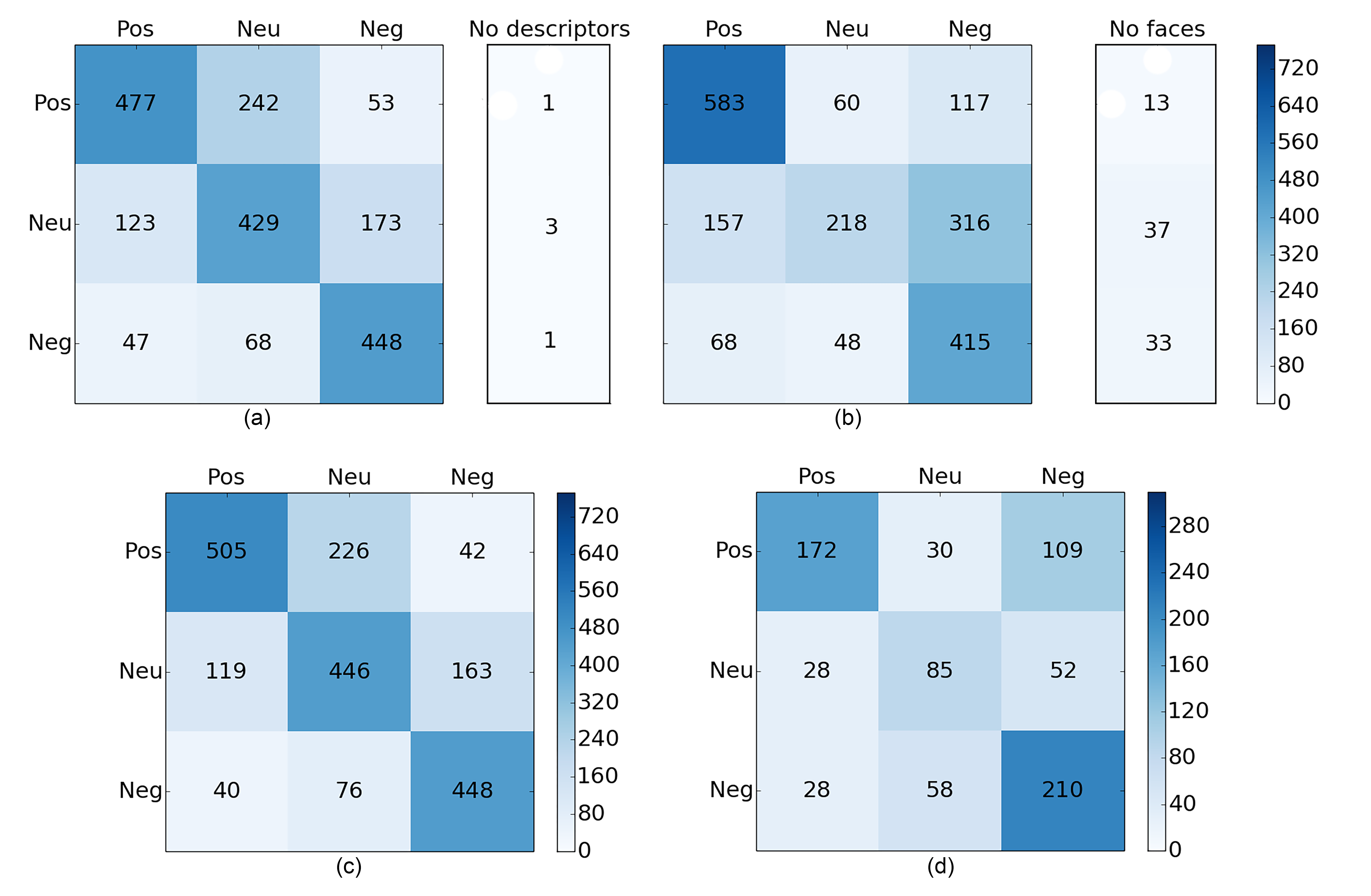}
\caption{Confusion matrices for (a) BN on validation set; (b) CNN on validation set; (c) complete system on validation set; (d) complete system on test set. Color scale is based on the accuracy value.
\label{fig:conf_matrices}}
\end{figure}

Considering the confusion matrices for the three methods, reported in figure  \ref{fig:conf_matrices}, we can see how the integration of BN and CNN leads to the best performance. The darker cells are on the main diagonal, meaning that the system can associate unknown input features to correct labels.

%SECTION: Conclusion and Future Work
\section{Conclusions and Future Work}
In this article we investigated the use of deep CNNs and Bayesian classifiers for group emotion recognition in the wild. Our method uses an approach which takes into account both top-down and bottom-up methods. The top-down module estimates the group emotion based on scene descriptors, which are integrated in a BN. On the other hand the bottom-up module identifies human faces in the picture and returns an average emotion estimation. The output of the bottom-up module is then redirected to the BN in the higher layer and considered as a dependent node. The method has been tested on the EmotiW'17 challenge dataset, obtaining an accuracy of $67.75\%$ on the validation set and $64.68\%$ on the test set, and achieving a significant improvement over the baseline performance \cite{dhall2015more}.

Future work should focus on different aspects which may have an important role on the accuracy. A more sophisticated approach for integrating the output of the CNN for each detected face should be taken into account. For example in \cite{dhall2015automatic} and  \cite{vonikakis2016group}, a weighted average based on the size of the face is used in order to calculate an overall score. This method is reasonable since it gives a lower weight to faces which are on the background, and which may carry less information.
In order to further improve the classification accuracy obtained through the CNN it is possible to use a divide-and-conquer strategy. Instead of relying on a single CNN to estimate the three categories, it may be possible to split the classifier in three sub-networks which are specialized in the identification of a single emotion. Such an approach showed major improvements in different tasks \cite{patacchiola2017head, cho1995combining, collins1988application, nguyen2004multiple}.

Another critical point is the integration of the estimates made by the two modules. In this work we empirically found that redirecting the output of the bottom-up module to the BN in the top layer leads to a slight improvement. However, other methods, for instance bagging \cite{breiman1996bagging}, should be considered in this delicate phase.
In conclusion, further research is needed in order to understand which approach may lead to higher performances.

\begin{acks}
We gratefully acknowledge the support of NVIDIA Corporation with the donation of the Tesla K40 GPU used for this research.
\end{acks}

\bibliography{sample-bibliography} 

\end{document}